# HYBRID AUTOENCODER-ISOLATION FOREST APPROACH FOR TIME SERIES ANOMALY DETECTION IN C70XP CYCLOTRON OPERATION DATA AT ARRONAX

F. Basbous[1], F. Poirier[2], F. Haddad[1,2], ARRONAX, Saint Herblain, France
D. Mateus, Nantes University, École Centrale Nantes, LS2N, UMR 6004, France
[1]also at Nantes Université, Nantes, France, [2]also at CNRS, France

*Abstract*

The Interest Public Group ARRONAX's C70XP cyclotron, used for radioisotope production for medical and research applications, relies on complex and costly systems that are prone to failures, leading to operational disruptions. In this context, this study aims to develop a machine learning-based method for early anomaly detection, from sensor measurements over a temporal window, to enhance system performance. One of the most widely recognized methods for anomaly detection is Isolation Forest (IF), known for its effectiveness and scalability. However, its reliance on axis-parallel splits limits its ability to detect subtle anomalies, especially those occurring near the mean of normal data. This study proposes a hybrid approach that combines a fully connected Autoencoder (AE) with IF to enhance the detection of subtle anomalies. In particular, the Mean Cubic Error (MCE) of the sensor data reconstructed by the AE is used as input to the IF model. Validated on proton beam intensity time series data, the proposed method demonstrates a clear improvement in detection performance, as confirmed by the experimental results.

## INTRODUCTION

Early detection of faults in safety-critical industrial systems is essential to prevent failures, irreversible losses, and costly repairs [1]. In this context, anomaly detection has become a key subdomain in the field of particle accelerators, focusing on identifying deviations from normal system behavior [2-4]. At ARRONAX [3], where beam stability is crucial, machine learning algorithms such as Density Based Spatial Clustering of Applications with Noise (DBSCAN) and Isolation Forest (IF) are used to explore their ability to highlight anomalies during long runs, while at SuperKEKB [5] residual machine learning models are applied to monitor vacuum and beam signals to prevent beam loss, and at Jefferson Lab [6] Principal Component Analysis (PCA) is used to detect abnormal behaviour in the superconducting RF cavities of CEBAF. Among the various algorithms proposed for this purpose, IF [7] represents a particularly relevant approach. As one of the first methods specifically developed for anomaly detection, IF has become a benchmark due to its effectiveness and scalability, and it has been widely applied to diverse real-world problems [8-11]. Despite these strengths, IF also shows certain limitations. To address them, three main strategies have been proposed in the literature [12].

The first strategy focuses on method improvements, where the algorithm itself is modified. This includes enhancing the tree construction process, such as using random projections to overcome axis-parallel split limitations [13], and refining the anomaly scoring stage through path-weighted scores [14], fuzzy scoring [15], or Deep Isolation Forest [16], which operates in a neural latent space with a modified scoring scheme.

The second strategy relies on data pre-processing before applying IF. Puggini and McLoone [17] reduce the bias caused by correlated features using feature selection and dimensionality reduction, while Chen et al. [18] handle multi-context data by clustering with Gaussian Mixture Models (GMM) and then applying IF within each cluster.

The third strategy involves post-processing the outputs of IF. Aminanto et al. [19] reduce false positives by filtering IF scores through a stacked autoencoder, whereas Alsini et al. [20] enhance the detection of local anomalies by combining IF with a sliding-window refinement followed by the Local Outlier Factor (LOF).

To the best of our knowledge, most studies have focused on modifying the algorithm itself, while the crucial role of feature space transformation has received considerably less attention. Yet, the way data are represented strongly affects the ability of IF to separate normal and anomalous instances. Although IF performs well in detecting global anomalies, it often fails to capture local or subtle ones, which in our case are critical as they may indicate early component degradation. As a contribution, we propose a hybrid AE-IF model, where an Autoencoder (AE) replaces the raw intensity inputs with their reconstruction errors, thus projecting the data into a reconstruction-error feature space where subtle anomalies are amplified, allowing IF to detect them effectively.

## METHOD

To illustrate the overall workflow, Fig. 1 compares the standard IF with the proposed AE-IF methodology. In the standard approach, the temporal beam intensity signal $x(t)$ is segmented into time windows, and IF directly assigns anomaly scores and binary labels. In the proposed method, an AE first reconstructs each window and computes the Mean Cubic Error (MCE) as a new feature space. IF is then applied to these error features. Since the AE is trained only on normal data, anomalous sequences produce higher reconstruction errors, making them more distinguishable and easier for IF to detect.

### Anomaly Types

From discussions with the ARRONAX cyclotron experts, two types of anomalies were identified, as illustrated in Fig. 2. Global anomalies, such as b, c, and d, correspond to clear drops or rises in the signal where at least one observation lies outside an expert defined interval $[S_{low}, S_{high}]$. These deviations are easily visible at the global scale. In contrast, subtle local anomalies are more difficult to perceive, since all values remain within $[S_{low}, S_{high}]$, but the internal variability of the window is high and exceeds a parameter $\alpha$. An example is shown in the zoomed view, where anomaly (a), hardly perceptible at the wide scale, exhibits significant fluctuations within the thresholds.

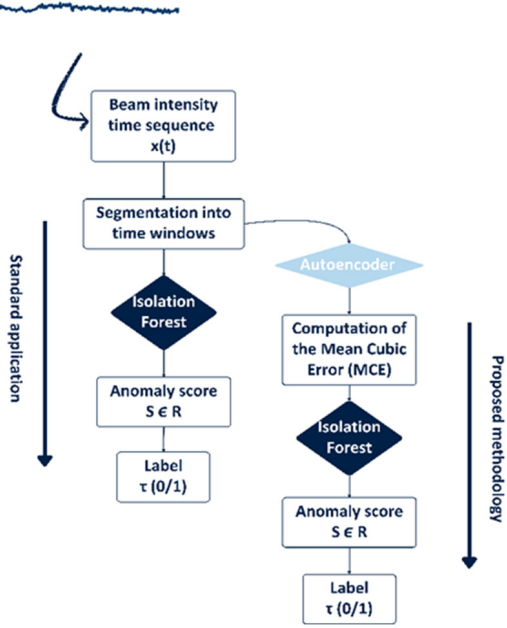

Figure 1: Comparaison between the standard IF application and the proposed AE-IF methodology.

## Isolation Forest

IF, introduced by Liu et al. [7], is an unsupervised method dedicated to anomaly detection. Its principle is based on the idea that abnormal observations are rare and different, and can therefore be isolated more easily than normal points [12]. The algorithm constructs a set of binary trees from random subsamples of the data [7]. At each node of a tree, a feature and a threshold value are chosen randomly to partition the space. Thus, outliers, located on the periphery of the data, are isolated in a small number of divisions, corresponding to a shorter path depth. Conversely, normal points require more partitions before being isolated. The anomaly score $s(x, n)$ of an instance $x$ is calculated using Eq. (1).

$$s(x,n) = 2^{-\frac{E[h(x)]}{c(n)}}, \quad (1)$$

where $E[h(x)]$ is the average path length to isolate $x$ across all isolation trees, and $c(n)$, defined in Eq. (2), is the expected average path length of unsuccessful searches in a binary search tree built from $n$ data points.

$$c(n) = 2H(n-1) - \frac{2(n-1)}{n}. \quad (2)$$

$$H(m) = ln(m) + \beta, \quad (3)$$

in this equation, $H(m)$ is the harmonic number and $\beta$ is the Euler constant. A data point is considered anomalous if its anomaly score exceeds the contamination rate $\tau$, which represents the expected proportion of anomalies in the dataset.

## Autoencoder

An AE [21] is an unsupervised neural network composed of three elements. An encoder that reduces the input into a latent representation, a bottleneck that holds this compressed representation, and a decoder that aims to rebuild the input from the latent code. The training process is guided by minimizing a reconstruction loss, which measures the difference between the input and the output. The Mean Squared Error (MSE) is one of the most commonly used reconstruction losses.

$$L_{MSE}(E, D) = \frac{1}{N}\sum_{i=1}^{N}\left(x_i - D(E(x_i))\right)^2, \quad (4)$$

where $E(.)$ represents the encoder, $D(.)$ the decoder, and $N$ the number of training samples. In this work, we used a simple fully connected AE with one encoder layer, and a symmetric decoder.

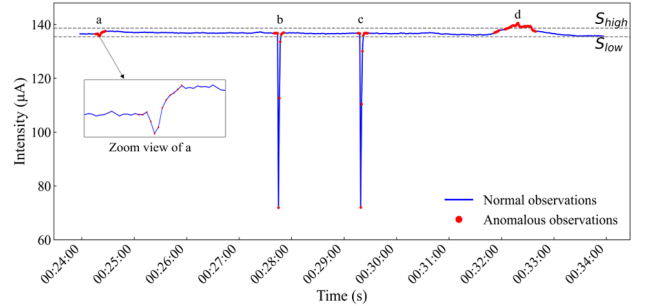

Figure 2: Examples of beam intensity anomalies (a-d), from local variations to breakdowns.

## Training Phase

The training set consists of $N$ vectors of dimension $a$, denoted $X = \{x_i\}_{i=1}^{N}$, standardized to have zero mean and unit variance. In the first step, only the normal samples of the training set are used to train the AE, which is optimized with the MSE loss to capture the intrinsic structure of normal beam intensity patterns. Once trained, the AE is applied to the entire standardized training set (normal and anomalous samples). For each input vector $x_i$, the reconstruction $\hat{x}_i$ is compared to the original using the Mean Cubic Error (MCE), defined in Eq. (5)

$$MCE(x_i, \hat{x}_i) = \frac{1}{a}\sum_{j=1}^{a}\left(x_{ij} - \hat{x}_{ij}\right)^3, \quad (5)$$

where $a$ is the input dimension. This transformation yields a new feature space, referred to as the MCE set (normal and anomalous samples), which is then used to train the IF for distinguishing between normal and anomalous data.

## Testing Phase – Anomaly Detection

During testing, each input x from the test set is first standardized using the transformation learned during training and reconstructed by the trained AE. The MCE is then computed to measure the reconstruction discrepancy.

Normal inputs are typically reconstructed with low error, while anomalous inputs yield higher errors. The resulting MCE values are processed by the trained IF, which assigns an anomaly score s(MCE(x)). The decision function used to produce the final prediction is defined in Eq. (6).

$$f(x) = s\big(MCE(x)\big) - \tau, \qquad (6)$$

where samples are classified as anomalies if $f(x) < 0$ and as normal otherwise. The threshold $\tau$ is the contamination rate previously defined in the Isolation Forest section.

Table 1: Average Detection Performance on the Test Data

| Method | Recall | Precision | F1-score | AUC-PR |
|---|---|---|---|---|
| IF | 0.47 | 0.45 | 0.41 | 0.49 |
| PCA-IF | 0.67 | 0.63 | 0.61 | 0.74 |
| AE-IF | **0.81** | **0.94** | **0.87** | **0.87** |

## RESULTS AND DISCUSSION

Twenty-five univariate time series of proton beam intensity (7-10 days each) were collected, resampled at 1 Hz [3], and segmented into non-overlapping windows of length k=6. Model development was performed on 60% of the runs (15 runs) using 10-fold cross-validation, where in each fold 10 runs were used for training and 5 for validation. The remaining 40% (10 unseen runs) were reserved for final testing.

The performance of the proposed AE-IF model is compared with two reference approaches, the standard IF applied to the raw intensity data and the PCA-IF variant, where PCA is used to reduce the dimensionality of the intensity features before applying IF. In the next part of this section, we present and discuss both the quantitative and qualitative results obtained for the compared models.

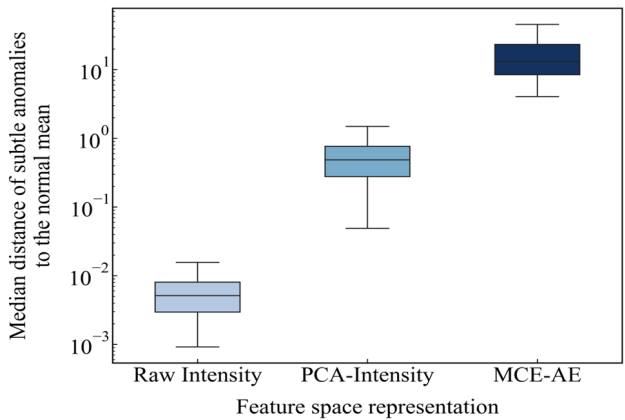

Figure 3: Subtle anomaly separability across feature spaces.

### Relative Deviations Across Feature Spaces

To assess the impact of feature space representation on anomaly separability, the relative distances between local anomalous instances and the mean of normal data were computed in three distinct spaces, the raw intensity space, the PCA transformed space, and the reconstruction error space produced by the AE (MCE-AE).

As shown in Fig. 3, in the raw intensity space, subtle anomalies remain very close to the normal mean, with a median distance below $10^{-2}$, which makes them almost indistinguishable from normal data. The PCA transformation increases the separation, although some anomalies still overlap with the normal distribution, with a median distance remaining below 1. In contrast, in the MCE-AE space, anomalies are much more distant from the normal mean, with a median distance around 13. This illustrates how the choice of feature space can amplify subtle anomalies, transforming them from overlapping patterns in the raw intensity space into clearly visible outliers in the reconstruction error space.

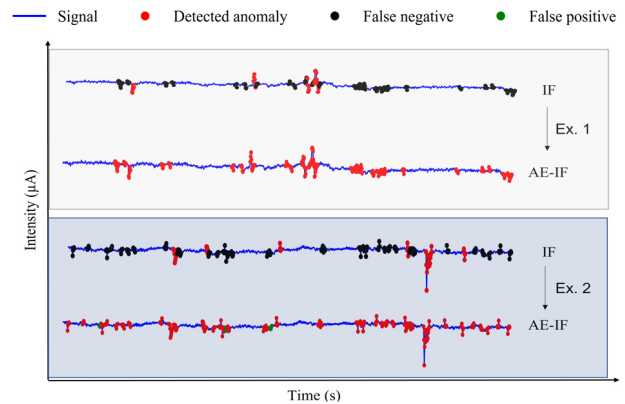

Figure 4: Detection comparison between IF and AE-IF.

### Detection Performance

The detection performance of IF applied to raw intensities (IF), PCA-transformed features (PCA-IF), and autoencoder reconstruction errors (AE-IF) was evaluated on the final test set using recall, precision, F1-score, and Area Under the Precision-Recall Curve (AUC-PR).

The results, summarized in Table 1, show that AE-IF consistently outperforms the other two approaches. It combines both high recall (0.81) and high precision (0.94), indicating its ability to detect a larger fraction of anomalies while maintaining a low false alarm rate. IF applied to raw intensities remains the least effective, while PCA-IF achieves intermediate performance by partially improving anomaly separability. The superiority of AE-IF is confirmed by the F1-score (0.87), which reflects the balance between recall and precision, and by AUC-PR (0.87), showing stable performance across thresholds. Figure 4 illustrates this behavior, where subtle anomalies that remain undetected in the raw intensity space are correctly identified in the AE-IF representation, while normal instances continue to be classified correctly.

## CONCLUSION AND PERSPECTIVES

The proposed AE-IF model outperforms the two other IF variants applied to the raw and PCA-transformed intensity spaces, effectively detecting both subtle and global anomalies. These results emphasize the crucial role of feature space representation prior to any algorithmic modification.

Beyond detection performance, this approach opens perspectives for operational monitoring, the development of irradiation stabilization methodologies, reduction of irradiation time in radionuclide production and potential extensions to flash-therapy irradiation analysis at Arronax.

## ACKNOWLEDGEMENTS

The Arronax cyclotron is supported by the CNRS, Inserm, INCa, Nantes Université, the Regional Council of Pays de la Loire, local authorities, the French government, and the European Union. This work has been supported in part by the French National Research Agency (ANR) "France 2030 investment plan" under the references I-SITE NExT (ANR-16-IDEX-0007), and Labex DHOLMEN, by financial support from the Pays de la Loire Region and by a grant from INCa-DGOS-INSERM-ITMO Cancer_18011 (SIRIC ILIAD).